\documentclass[preprint,12pt]{elsarticle}

\usepackage{amssymb}
\usepackage{amsmath}
\usepackage{multirow}
\journal{Neural Networks}

\begin{document}

\begin{frontmatter}



\title{Geometric Transformation-Embedded Mamba for Learned Video Compression}


%\author{Hao Wei, Yanhui Zhou, Jingwen Jiang, Chenyang Ge} %% Author name
\author[label1]{Hao Wei}
\author[label2]{Yanhui Zhou}
\author[label1]{Chenyang Ge}
\affiliation[label1]{organization={State Key Laboratory of Human-Machine Hybrid Augmented Intelligence, Institute of Artificial Intelligence and Robotics, Xi'an Jiaotong University},
             addressline={No.28, West Xianning Road}, 
             city={Xi'an},
             postcode={710049},
             state={Shaanxi},
             country={China}}
\affiliation[label2]{organization={School of Information and Communications Engineering, Xi'an Jiaotong University},
             addressline={No.28, West Xianning Road}, 
             city={Xi'an},
             postcode={710049},
             state={Shaanxi},
             country={China}}

%% Abstract
\begin{abstract}
%
%Perceptual video compression aims to reduce bitrates while maintaining high visual perceptual quality, which is essential for bandwidth- or storage-limited applications.
Although learned video compression methods have exhibited outstanding performance,
most of them typically follow a hybrid coding paradigm that requires explicit motion estimation and compensation, resulting in a complex solution for video compression. %leading to high model complexity.   
In contrast, we introduce a streamlined yet effective video compression framework founded on a direct transform strategy, i.e., nonlinear transform, quantization, and entropy coding.
We first develop a cascaded Mamba module (CMM) with different embedded geometric transformations to effectively explore both long-range spatial and temporal dependencies. 
To improve local spatial representation, we introduce a locality refinement feed-forward network (LRFFN) that incorporates a hybrid convolution block based on difference convolutions. 
We integrate the proposed CMM and LRFFN into the encoder and decoder of our compression framework. 
Moreover, we present a conditional channel-wise entropy model that effectively utilizes conditional temporal priors to accurately estimate the probability distributions of current latent features. 
Extensive experiments demonstrate that our method outperforms state-of-the-art video compression approaches in terms of perceptual quality and temporal consistency under low-bitrate constraints. Our source codes and models will be available at https://github.com/cshw2021/GTEM-LVC.
\end{abstract}

%% Keywords
\begin{keyword}


Learned video compression \sep cascaded Mamba \sep geometric transformation \sep difference convolution \sep conditional entropy model
\end{keyword}

\end{frontmatter}



\section{Introduction}
\label{Introduction}
As demand for high-quality video content continues to grow and its applications expand, it has become a cornerstone of the digital landscape. Consequently, methods of compressing videos for storage and transmission have become critical.

\begin{figure*}[htbp]
\centering
\includegraphics[width=1.0\textwidth]{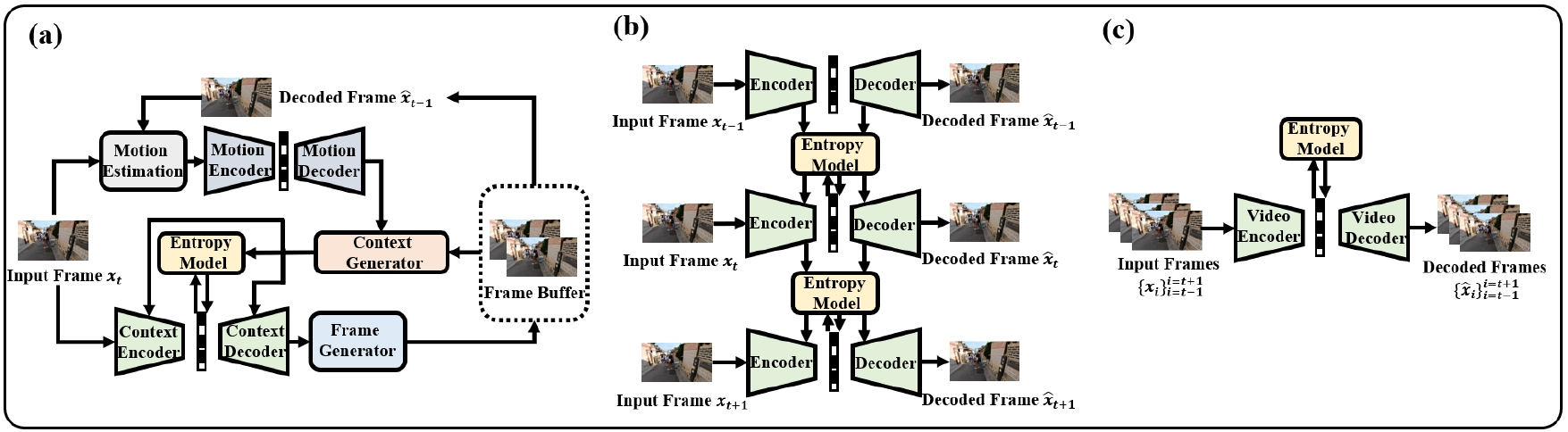}\vspace{-3mm}
\caption{Architectural comparisons with existing video compression methods: (a) shows a hybrid coding framework; (b) is a transform-based coding method that is both frame-independent and latent-dependent; (c) illustrates the proposed coding approach that is frame-dependent and latent-dependent.} 
\label{arch_com}
\end{figure*}

While image compression primarily focuses on exploiting spatial dependencies to reduce redundancy~\cite{ICISP_NN2025}, video compression is more complex as it needs to capture both spatial and temporal dependencies. Most existing methods adopt a learned hybrid coding framework derived from traditional hybrid video codecs \cite{DVC_TPAMI2020, DCVC_NeurIPS2021, DCVC_HEM_ACMMM2022}. As shown in Fig. \ref{arch_com}(a), the output frame (after decompression) is typically reconstructed using motion estimation and compensation networks. The motion vector and residual, calculated either explicitly or conditionally, are then compressed. However, these methods require complex solutions, such as residual coding, motion coding, and motion estimation and compensation. 
%Distributed coding methods further reduce encoding complexity while increasing decoder complexity, as these methods require side information to be generated at the decoder. This is typically achieved using optical flow-based frame interpolation technology \cite{Distributed_DVC_ICME2023} (Fig. \ref{arch_com}(b)).
%

Recently, transform-based video compression has gained increasing attention as an alternative to motion-compensated approaches, as it avoids explicit motion estimation and compensation \cite{3DAutoencoder_ICCV2019, STA_SIPS_2020, CEC_ECCV2020, VCT_NeurIPS2022}. Early works such as \cite{3DAutoencoder_ICCV2019, STA_SIPS_2020} employ 3D convolutions in the nonlinear transform to jointly model spatial and temporal information. However, the inherently local receptive fields of 3D convolutions limit their ability to capture long-range dependencies across spatial and temporal dimensions.
To address this limitation, Liu et al. \cite{CEC_ECCV2020} encode video frames independently using 2D convolutional neural networks and exploit temporal correlations through conditional entropy modeling at both the encoder and decoder (see Fig. \ref{arch_com}(b)). Building upon this framework, VCT \cite{VCT_NeurIPS2022} introduces a transformer-based entropy model conditioned on the latent representations of previously encoded frames, enabling more expressive temporal dependency modeling. Nevertheless, conditioning solely on past latent features is insufficient to fully characterize complex temporal dependencies, leading to suboptimal compression performance.

In this paper, we propose a simple and effective transform-based video compression method that leverages long-range spatial and temporal dependencies for redundancy reduction (see Fig. \ref{arch_com}(c)). Motivated by the success of state space models in capturing global contexts, we first develop a cascaded Mamba module that scans video frames in the forward and backward directions with respect to both spatial and temporal dimensions by embedding different geometric transformations. Moreover, to capture local spatial information, we design a locality refinement feed-forward network that incorporates difference convolutions, which focus on capturing fine-grained details. 
We formulate them as core components of the encoder–decoder architecture, addressing the limitations of existing 2D and 3D convolution-based approaches.
In addition, we propose a conditional channel-wise entropy model that leverages pseudo-latent features of the current frame as auxiliary conditions instead of relying solely on latent features from the previously decoded frames.
%Specifically, given two previously decoded latent features, we propose a predictive motion alignment module that generates pseudo aligned features. These features are then fused with the previously decoded ones through a condition generation network. The resulting fused features serve as condition priors for entropy estimation, greatly improving compression performance. 

The main contributions are summarized as follows:
\begin{itemize}
    \item We present a frame- and latent-dependent transform-based video compression method that achieves competitive perceptual quality and temporal coherence across bitrates.
    \item A cascaded Mamba module with different geometric transformations is developed to capture spatial and temporal non-local dependencies across video frames. In addition, we design a locality refinement feed-forward network to model local dependencies more effectively.
    \item We develop a conditional channel-wise entropy model that utilizes both previously decoded latent features and pseudo latent features of the current frame as conditions to assist in encoding the current frame.
\end{itemize}

\section{Related Work}
\label{Related_work}
\subsection{Neural Video Compression}
Learned video compression has become a sought-after research topic in recent years. Existing learned video compression methods are generally categorized into three groups: hybrid video coding \cite{DVC_TPAMI2020, FVC_CVPR2021, SSF_CVPR_2020, NVC_TCSVT2020, SaEC_CVPR2023, TCVC_TIP2023, DCVC_NeurIPS2021, DCVC_TCM_TMM2022, DCVC_HEM_ACMMM2022, DCVC_DC_CVPR2023, DCVC_MIP_CVPR2023, DCVC_FM_CVPR2024, DCVC_LCG_ECCV2025}, distributed video coding \cite{Distributed_DVC_ICME2023}, and transform-based video coding \cite{3DAutoencoder_ICCV2019, SPA_SiPS2020, CEC_ECCV2020, VCT_NeurIPS2022, DHVC_AAAI2024}.

\textbf{Hybrid Video Coding.}
Lu et al. \cite{DVC_TPAMI2020} introduce the first end-to-end deep video compression model that adheres to traditional video compression approaches based on the hybrid coding framework, encompassing motion estimation and compensation, residual compression, motion compression, and bitrate estimation. To enhance motion compensation, FVC \cite{FVC_CVPR2021} shifts motion-related operations from pixel space to feature space, employing deformable convolution \cite{DeformConv_ICCV2017} for better alignment. Liu et al. \cite{NVC_TCSVT2020} develop a multiscale motion compensation network that uses multiscale compressed flows for alignment in a coarse-to-fine fashion. In \cite{HMC_TCSVT2022}, a hybrid motion compensation method is proposed, adaptively combining both pixel-level and feature-level compensation. Gao et al. \cite{SPME_ACMMM2022} introduce auxiliary motion estimation to address inaccuracies in motion estimation caused by previously decoded frames.

Another approach within the hybrid video coding framework focuses on conditional coding rather than residual compression. Li et al. \cite{DCVC_NeurIPS2021} develop a context refinement module that generates conditions to assist in encoding, decoding, and entropy modeling. In \cite{DCVC_TCM_TMM2022}, a temporal context mining module is proposed to learn multiscale temporal contexts, which are used as priors for compression. DCVC-HEM \cite{DCVC_HEM_ACMMM2022} introduces a hybrid entropy model that incorporates both spatial and temporal contexts. In \cite{DCVC_DC_CVPR2023}, Li et al. enhance the context diversity by employing a group-based offset diversity and a quadtree-based partition strategy. Qi et al. \cite{DCVC_MIP_CVPR2023} propose a hybrid context generation module that creates both multiscale contexts and motion conditions, which help propagate motion information and reduce alignment errors. The DCVC-LCG framework \cite{DCVC_LCG_ECCV2025} integrates long-term context with short-term reference features to refine feature propagation and improve reconstruction quality. 

\textbf{Distributed Video Coding.}
Unlike hybrid predictive coding methods, distributed video coding shifts the motion-related computational burden from the encoder to the decoder. In \cite{Distributed_DVC_ICME2023}, Zhang et al. propose a side information generation network to capture temporal correlations at the decoder. However, generating side information through interpolation of frames at arbitrary time steps requires significant computation cost and can lead to suboptimal compression performance, particularly when optical flow estimation is inaccurate due to large motion or occlusion.

\textbf{Transform-based Video Coding.}
Transform-based video coding typically follows a pipeline of transformation, quantization, and entropy coding. In \cite{3DAutoencoder_ICCV2019}, Habibian et al. utilize a 3D autoencoder based on 3D convolutions to encode a group of frames. Liu et al. \cite{CEC_ECCV2020} propose encoding each frame independently using an image compression method and then explore temporal correlations in the latent space through conditional entropy modeling. Additionally, VCT \cite{VCT_NeurIPS2022} leverages the Transformer architecture to capture temporal dependencies in the latent representations of successive frames. To further enhance compression performance, Lu et al. \cite{DHVC_AAAI2024} exploit the multiscale properties of frames, using multi-scale latent features as priors for conditional probability estimation.

\subsection{Perceptual Video Compression}
While existing neural video compression methods perform well on pixel-level distortion metrics, they often produce decoded frames that appear overly smooth and lacking in detail at low bitrates. This limitation has motivated researchers to explore perceptual video compression techniques \cite{RDP_ICML_2019}. Leveraging the powerful generative capabilities of generative adversarial networks (GAN) \cite{GAN_NIPS_2014}, Mentzer et al. \cite{NVCGAN_ECCV_2022} propose the first GAN-based video compression model, in which the discriminator guides the compression network to generate videos with realistic details through adversarial training. Zhang et al. extend the DVC \cite{DVC_TPAMI2020} framework to a perceptual version by employing a discriminator and mixed loss function \cite{DVC_P_VCIP_2021}. In \cite{PLVC_IJCAI_2022}, Yang et al. design a recurrent conditional GAN that leverages spatial-temporal features as conditions for video compression. To enhance the reconstruction of newly emerged regions, a confidence-based feature reconstruction method is proposed in \cite{HVFVC_ACMMM_2023}. Du et al. \cite{CGVCT_2024} introduce contextual coding into a GAN-based video compression model to enhance coding efficiency. Additionally, Ma et al. \cite{DiffVC_2025} leverage a diffusion model for high-quality video generation, using the previously decoded frame and the reconstructed latent representation of the current frame as guidance.

\subsection{Vision Mamba and Its Application}
Recently, state space models such as Mamba have shown promising applications in various vision tasks due to their efficiency in long-range modeling \cite{VisionMamba, VMamba, VideoMamba_ECCV, MambaIR_ECCV}. In \cite{MambaVC}, Qin et al. propose a Mamba-based video compression framework built on SSF \cite{SSF_CVPR_2020}. Specifically, they replace the CNN-based transforms with a visual state space block that implements a four-way selective scanning along the spatial dimension \cite{VMamba}. In our work, we extend this concept by introducing bidirectional 3D scanning. The proposed approach captures the long-range dependencies across both spatial and temporal dimensions in a cascaded manner.
\section{Methodology}
\label{Methodology}
\subsection{Overall Framework}
\begin{figure*}[htbp]
\centering
\includegraphics[width=1.0\textwidth]{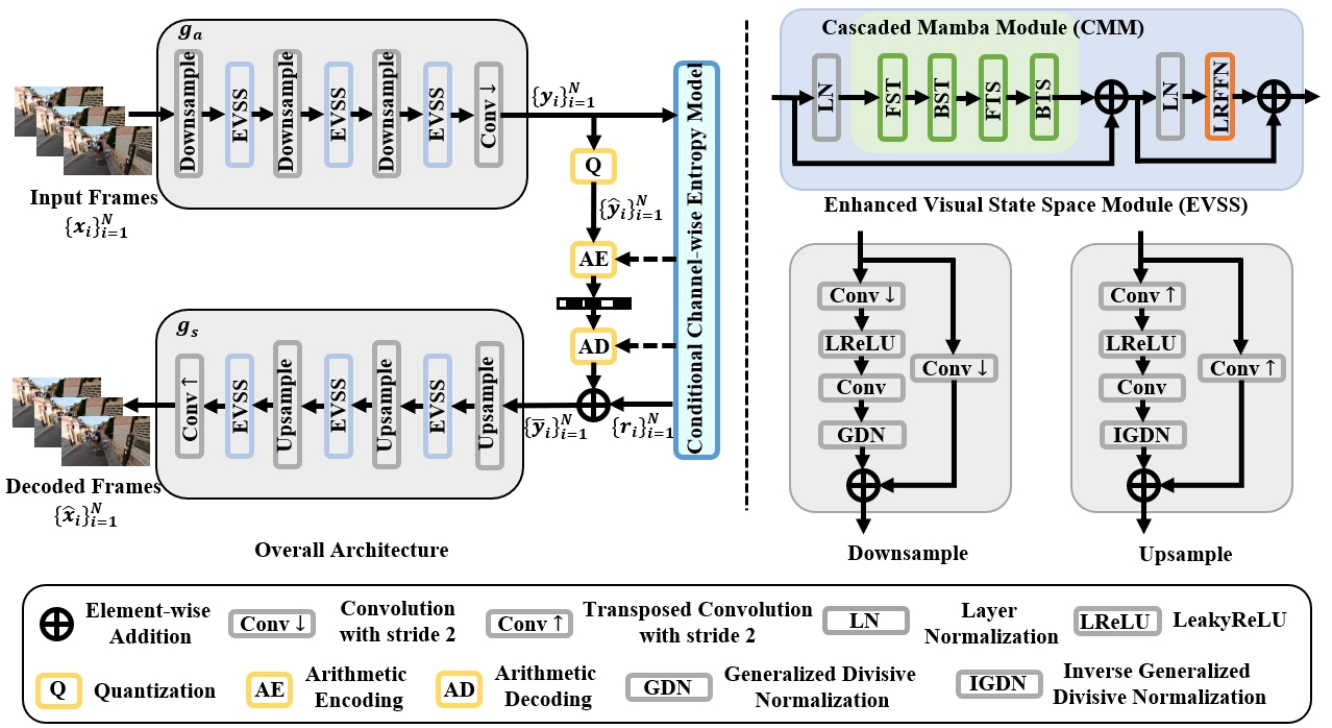}\vspace{-3mm}
\caption{The overall architecture of the proposed video compression framework. To capture both long-range spatio-temporal dependencies and local dependencies, we develop the cascaded Mamba module (CMM) for global modeling and locality refinement feed-forward network (LRFFN) for local modeling. The CMM is implemented by sequentially traversing video features in four directional orders, namely forward spatio-temporal (FST), backward spatio-temporal (BST), forward temporal–spatial (FTS), and backward temporal–spatial (BTS).} 
\label{arch}
\end{figure*}
The overall pipeline of the proposed transform-based video compression framework is illustrated in Fig. \ref{arch}. Given a video sequence $\left \{ x_{i}  \right \} _{i=1}^{N}$ consisting of $N$ frames, the encoder $g_a$ maps these consecutive frames to their latent representations $\left \{ \text{y}_{i} \right \}_{i=1}^{N}$. The latent representations are subsequently discretized to $\left \{ \hat{\text{y}}_{i} \right \}_{i=1}^{N}$  by a quantization operation and losslessly encoded and decoded via an arithmetic codec. In parallel, the proposed conditional channel-wise entropy model is used to estimate the latent residuals $\left \{ r_{i} \right \}_{i=1}^{N}$ and the distribution parameters of the latent features. Subsequently, the refined latents $\left \{ \bar{\text{y}}_{i} \right \}_{i=1}^{N}$ are obtained by summing up the latent residuals $\left \{ r_{i} \right \}_{i=1}^{N}$ and latent features $\left \{ \hat{\text{y}}_{i} \right \}_{i=1}^{N}$. Finally, the video frames $\left \{ \hat{x}_{i}  \right \} _{i=1}^{N}$ are reconstructed by feeding the refined latents to the decoder $g_s$.

For details of the proposed framework, we design the enhanced visual state space (EVSS) module as the core component of both the encoder and decoder. The EVSS consists of the cascaded Mamba module and the locality refinement feed-forward network, which are described in Sections \ref{CMM} and \ref{LRFFN}, respectively. Further details of the conditional channel-wise entropy model are provided in Section \ref{ccem}.

\subsection{Cascaded Mamba Module}
\label{CMM}
The vanilla Mamba block, originally designed for modeling 1D sequential data, struggles with vision tasks that require spatially informed processing \cite{Mamba_2023}. 
Several multi-directional scanning strategies have been proposed to compensate for the loss of spatial dependencies when flattening an image into a 1D sequence \cite{VisionMamba, VMamba}. Unfortunately, performing repetitive scanning across multiple directions in parallel introduces significant computational overhead. 
\begin{figure}
\centering
\includegraphics[width=0.7\textwidth]{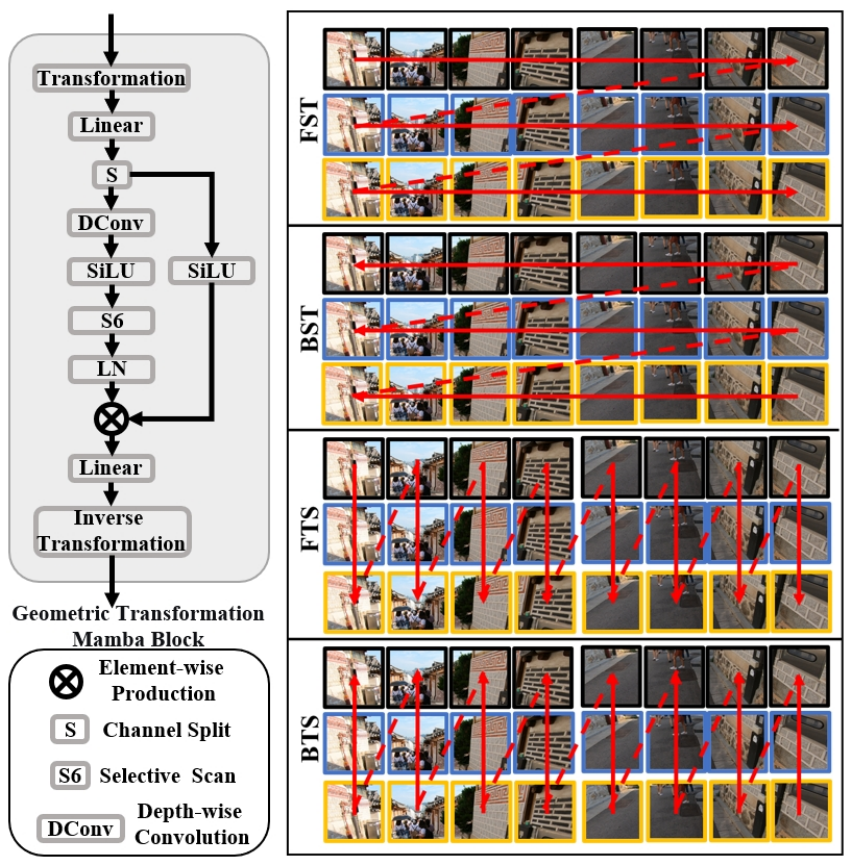}\vspace{-3mm}
\caption{Detailed architecture of geometric transformation Mamba block and different selective scanning methods. We first extract spatio-temporal video patches and then perform selective scanning along spatial and temporal dimension, depending on the priority assigned to each dimension.} 
\label{CMM_arch}
\end{figure}

To address this issue, we propose a geometric transformation Mamba block (GTMB) that incorporates a reversible transformation prior to scanning in a single direction, in contrast to the existing approach of repetitive scanning across multiple directions. The key idea is that applying a non-parametric, reversible transformation to video features indirectly modifies the selective scanning process. Reversibility is essential, as it allows us to recover the features back to their original spatio-temporal structure after scanning. The detailed architecture of GTMB is illustrated in Fig. \ref{CMM_arch}.
Given the input feature $\mathbf{X}$, the process can be formulated as follows: 
\begin{equation}
\label{mamba_eq}
\centering
\begin{split}
    \mathbf{X_1}, \mathbf{X_2} = \text{split}(\text{Linear}(\mathcal{T}(\mathbf{X}))), \quad
    \mathbf{\bar{X}_1} = \text{LN}(\text{S6}(\rho(\text{DConv}(\mathbf{X_1})))),\\
    \mathbf{\hat{X}} = \mathcal{T}^{-1}(\text{Linear}(\mathbf{\bar{X}_1}\otimes \rho(\mathbf{X_2}))),
\end{split}
\end{equation}
where $\text{Linear}(\cdot)$ denotes a learnable linear projection, $\text{DConv}(\cdot)$ represents a depth-wise convolution with a kernel size of 3$\times$3, $\text{LN}(\cdot)$ denotes the layer normalization, $\rho$ is the SiLU activation, $\text{split}(\cdot)$ splits the features evenly in the channel dimension, $\mathcal{T}$ and $\mathcal{T}^{-1}$ denote the transformation operation and its inverse version, which decide what scanning type to choose, respectively, and $\text{S6}$ denotes the selective scanning \cite{Mamba_2023}.

To effectively model long-range spatio-temporal dependencies and minimize redundancies for video compression, we use four selective scanning strategies that scan video features based on the priority assigned to the spatial or temporal dimension, as illustrated in Fig. \ref{CMM_arch}. 
Specifically, the forward spatial-temporal (FST) scan performs scanning frame by frame along the spatial dimension in a forward direction, with $\mathcal{T}$ in Eq.(\ref{mamba_eq}) being an identity transformation.
The backward spatial-temporal (BST) scan is realized by defining the transformation operation as a flipping operation over both the temporal and spatial dimensions of the video features.
In addition to scanning the spatial dimension first, we can also prioritize scanning the temporal dimension to capture temporal redundancy. The forward temporal-spatial (FTS) scan processes video patches at the same spatial location along the temporal dimension in a forward direction, with $\mathcal{T}$ defined as a transpose operation between the spatial and temporal dimensions. By using both flipping and transpose operations, we can achieve the backward temporal-spatial (BTS) scan. 

Based on the above scanning strategies, we propose the cascaded Mamba module (CMM), which is implemented with four bidirectional geometric Mamba blocks, each featuring a distinct transformation operation, to capture comprehensive global context across both spatial and temporal domains. The architecture of the CMM is shown in Fig. \ref{arch}. To be specific, given the normalized feature $\mathbf{F}$, the whole process can be formulated as:
\begin{equation}
    \mathbf{Y}=\mathcal{M}_{\text{BTS}}(\mathcal{M}_{\text{FTS}}(\mathcal{M}_{\text{BST}}(\mathcal{M}_{\text{FST}}(\mathbf{F})))),
\end{equation}
where $\mathbf{Y}$ is generated compact feature, $\mathcal{M}(\cdot)$ denotes the GTMB; the right subscript symbols $\left\{\text{BTS}, \text{FTS}, \text{BST}, \text{FST}\right\}$ represent different scanning approaches, i.e., backward temporal-spatial scan, forward temporal-spatial scan, backward spatial-temporal scan, and forward spatial-temporal scan, respectively.

\subsection{Locality Refinement Feed-forward Network}
\label{LRFFN}
\begin{figure}
\centering
\includegraphics[width=0.4\textwidth]{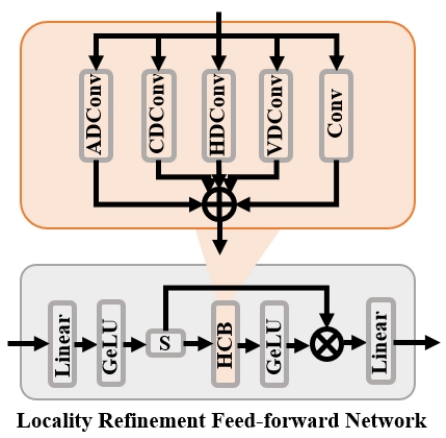}\vspace{-3mm}
\caption{Detailed architecture of the proposed locality refinement feed-forward network.} 
\label{lrffn_arch}
\end{figure}
The original feed-forward network (FFN) \cite{TBIC_ICLR2022, AICT_ICIP2023} processes features independently at each pixel location, facilitating feature refinement. However, the standard FFN struggles to capture local relationships between neighboring pixels, which are crucial for low-level vision tasks \cite{MKFFN_ECCV2025, AST_CVPR2024}. To address this limitation, existing methods either leverage block-wise Fast Fourier transform \cite{FTIC_ICLR2024} or introduce multiple depth-wise convolutions with varying kernel sizes \cite{MKFFN_ECCV2025} to better model local dependencies. In contrast to these methods, we propose a locality refinement feed-forward network (LRFFN) to focus on the local fine-grained features using a well-designed hybrid convolution block (HCB), complementary to the proposed STM that captures the global context. 

As shown in Fig. \ref{lrffn_arch}, the proposed LRFFN consists of two branches. One branch uses a hybrid convolution block to capture local relationships, which are activated by the GeLU activation and subsequently used to modulate the features from the other branch, reducing redundancy along the channel dimension. Overall, given an input feature $\mathbf{E}$, the process of LRFFN can be formulated as:
\begin{equation}
\label{LRFFN_equ}
    \begin{split}
        \mathbf{E_1}, \mathbf{E_2} = \text{split}(\gamma(\text{Linear}(\mathbf{E}))),\quad
        \mathbf{\hat{E}} = \text{Linear}(\gamma(\mathcal{H}(\mathbf{E_1}))\otimes\mathbf{E_2}),
    \end{split}
\end{equation}
where $\gamma$ represents the GeLU activation and $\mathcal{H}(\cdot)$ refers to the proposed HCB. The HCB is composed of five parallel convolution operations, including vertical difference convolution (VDConv), horizontal difference convolution (HDConv), angular difference convolution (ADConv), central difference convolution (CDConv), and vanilla convolution. Note that the vanilla convolution captures intensity information, while the difference convolutions \cite{CDC_CVPR2020} capture only variations between neighboring values, leading to a more compact representation that requires fewer bits.

%
%Before introducing the proposed HCB, we first review the concept of difference convolution. Difference convolution, as the name suggests, captures the differences between neighboring pixel values in the input \cite{CDC_CVPR2020}. It typically starts by calculating pixel differences before applying the standard kernel weights for convolution. This process is equivalent to convolving the input feature using the learnable weights shown in Fig. \ref{conv}. Note that difference convolution is able to capture only variations between neighboring values, leading to a more compact representation that requires fewer bits.

%Specifically, the proposed HCB comprises five distinct convolutions including vertical difference convolution (VDC), horizontal difference convolution (HDC), angular difference convolution (ADC), central difference convolution (CDC), and vanilla convolution (VC). The vanilla convolution captures intensity information, while the difference convolutions focus on extracting sparse information that results in a more compact representation. As shown in Fig. \ref{arch}, given the input feature $\mathbf{E_1}$ as used in Eq.(\ref{LRFFN_equ}), the whole process of the HCB can be formulated as:
%\begin{equation}
%\begin{split}
  %  \mathbf{\hat{E}_1} = VDC(\mathbf{E_1})\oplus HDC(\mathbf{E_1})
  %  \oplus ADC(\mathbf{E_1})\oplus CDC(\mathbf{E_1})\oplus VC(\mathbf{E_1}),
%\end{split}
%\end{equation}
%where $\oplus$ denotes the element-wise addition operation.
%

\subsection{Conditional Channel-wise Entropy Model}
\label{ccem}
The entropy model is designed to accurately capture the statistical distribution of latent representations of video frames, facilitating more efficient coding. 
Unlike existing entropy models that focus on single images \cite{FTIC_ICLR2024}, the proposed conditional entropy model improves this process by using previously decoded latent features as temporal priors, allowing it to estimate the distribution of the current frame's latent features. This approach effectively leverages temporal dependencies to enhance video compression.

\begin{figure}
\centering
\includegraphics[width=1.0\textwidth]{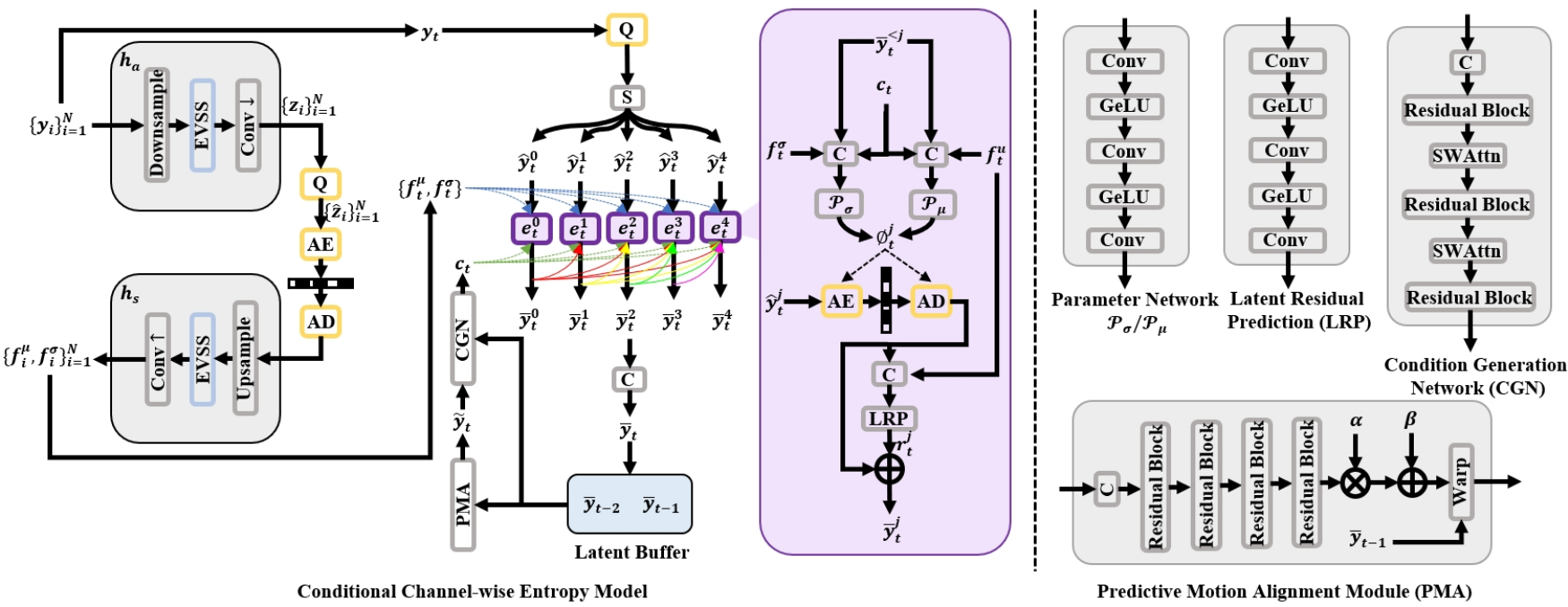}\vspace{-3mm}
\caption{Detailed architecture of the proposed conditional channel-wise entropy model.} 
\label{ccem_arch}
\end{figure}
Fig. \ref{ccem_arch} illustrates the architecture of the proposed conditional channel-wise entropy model. Given the sequential latent representations $\left \{ \text{y}_{i} \right \}_{i=1}^{N}$, the hyperprior networks $h_a$ and $h_s$ are employed to encode side information $\left \{ z_{i} \right \}_{i=1}^{N}$ and generate the parameters $\left \{ f_{i}^{\mu} \right \}_{i=1}^{N}$ and $\left \{ f_{i}^{\sigma} \right \}_{i=1}^{N}$. These features serve as inputs to the slice network $\left \{ e_{i} \right \}_{i=1}^{N}$, which produces the final refined latents $\left \{ \bar{\text{y}}_{i} \right \}_{i=1}^{N}$. 

Without loss of generality, we use the $t$-th frame as an illustrative example. Given the quantized latent $\hat{\text{y}}_t$, 
%and the condition feature $c_{t}$ which are generated by the proposed predictive motion alignment module and condition generation network, 
we first divide $\hat{\text{y}}_t$ into five slices, $\left \{\hat{\text{y}}_t^0, \hat{\text{y}}_t^1, \hat{\text{y}}_t^2, \hat{\text{y}}_t^3, \hat{\text{y}}_t^4 \right \}$, along the channel dimension. Assuming a Gaussian distribution for the $j$-th slice features, the slice network $e_t^j$ is employed to estimate the corresponding distribution parameters, namely the mean $\mu_t^j$ and scale $\sigma_t^j$, conditioned on the previously decoded slice features $\bar{\text{y}}_t^{<j}$ and hyper priors $\left \{ f_{t}^{\mu}, f_{t}^{\sigma} \right \}$. 
%The slice features are then processed by independent slice networks, in which the encoding of each slice is conditioned on information from previously encoded slices. 
%Taking the $j$-th slice as an example, 
we formulate the overall process as follows:
\begin{equation}
\begin{split}
    r_t^j, \phi_t^j = e
_t^j(f_t^{\mu}, f_t^{\sigma}, \bar{\text{y}}_t^{<j}, c_{t}), \quad
\bar{\text{y}}_t^j = r_t^j \oplus \hat{\text{y}}_t^j, \quad 0<=j<5,
\end{split}
\end{equation}
where $\phi_t^j=\left \{\mu_t^j,\sigma_t^j \right\}$ represents the probability parameters, $r_t^j$ denotes the latent residual, and $c_{t}$ denotes the conditional feature generated by the proposed predictive motion alignment module and condition generation network.
%Finally, the current refined latent $\hat{\text{y}}_t$ is obtained by aggregating $\left \{ \hat{\text{y}}_{t}^{i} \right \}_{i=0}^{4}$ through a channel-wise concatenation operation.
%

\textbf{Predictive Motion Alignment Module.}
\begin{figure}
\centering
\includegraphics[width=0.6\textwidth]{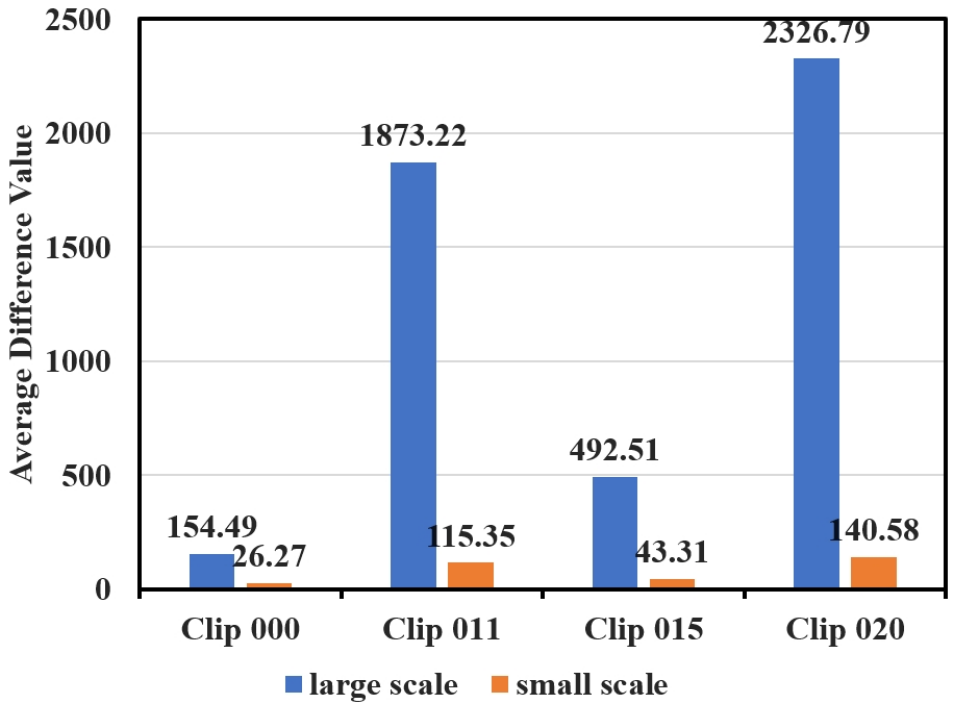}\vspace{-3mm}
\caption{The motion difference comparison among successive video frames.} 
\label{flowdiff}
\end{figure}
Unlike existing methods \cite{VCT_NeurIPS2022, DHVC_AAAI2024}, which only use previously decoded latents as conditions, we explore the motion between successive decoded latents to help current latent encoding. Given successive decoded latents, $\left \{\bar{\text{y}}_{t-2}, \bar{\text{y}}_{t-1} \right \}$, we first estimate the motion $f_{t-2\rightarrow t-1}$ between them using four residual blocks. This motion is then corrected using the learnable parameters $\left \{\alpha,\beta \right\}$ as follows:
\begin{equation}
    \bar{f}_{t-2 \rightarrow t-1} = \alpha\otimes f_{t-2 \rightarrow t-1}\oplus\beta,
\end{equation}
where $\bar{f}_{t-2 \rightarrow t-1}$ represents the rectified motion, which is subsequently used for feature alignment:
\begin{equation}
\label{warp}
    \tilde{\text{y}}_t = \mathcal{W}(\bar{f}_{t-2 \rightarrow t-1}, \bar{\text{y}}_{t-1}),
\end{equation}
where $\mathcal{W}$ is the warping operation and $\tilde{\text{y}}_t$ is the aligned feature corresponding to the current encoded latent $\text{y}_t$.
Note that in Eq.(\ref{warp}), we use the motion $\bar{f}_{t-2 \rightarrow t-1}$ for alignment instead of $f_{t-1 \rightarrow t}$ because the current latent $\text{y}_t$ is unavailable for motion estimation. A natural question that arises is whether $\bar{f}_{t-2 \rightarrow t-1}$ is a reasonable choice for alignment.   

The answer is YES. To verify its rationality, we first downscale the video frames by a factor of 16. We then use the pretrained SpyNet \cite{SpyNet_CVPR2017} to estimate the optical flow on both the video frames at their original resolution and on their downscaled version. Finally, we calculate the average motion difference between adjacent motions. As shown in Fig. \ref{flowdiff}, the motion difference calculated on the low-resolution videos is significantly smaller than that calculated on the high-resolution videos. This demonstrates that, in the latent space, the adjacent previous motion can serve as a pseudo ground truth for motion compensation when aligning the previous and current frames.

\textbf{Condition Generation Network.}
We propose a condition generation network (CGN) to generate conditional priors for the current frame encoding. Specifically, given the aligned feature $\tilde{\text{y}}_t$ and the decoded latents $\left \{\bar{\text{y}}_{t-2}, \bar{\text{y}}_{t-1} \right \}$, the CGN, which is implemented by alternately stacking residual block and swin-transformer attention block (SWAttn) \cite{TCM_CVPR_2023}, generates the fused condition $c_t$ by:
\begin{equation}
    c_t = \mathcal{C}(\tilde{\text{y}}_t,\bar{\text{y}}_{t-2}, \bar{\text{y}}_{t-1}),
\end{equation}
where $\mathcal{C}$ represents the CGN.

\subsection{Training Strategy}
We adopt a two-stage training strategy for the proposed video compression framework. In the first stage, we train the proposed video compression model via a rate-distortion loss as:
\begin{equation}
\label{rd_loss}
    L_{rd} = \sum_{i=1}^{N} R(\hat{\text{y}}_{i})+R(\hat{z}_i)+\lambda\cdot D(\hat{x}_i, x_i), 
\end{equation}
where $\lambda$ controls the rate-distortion tradeoff. $R(\hat{\text{y}}_i)$ and $R(\hat{z}_i)$ denote the bitrates of $i$-th frame's latent features $\hat{\text{y}}_i$ and quantized side information $\hat{z}_i$, respectively. $D(\hat{x}_i, x_i)$ represents the distortion calculated by mean squared error (MSE) loss. In the second stage, we add the perceptual loss and style loss for perception-oriented optimization as:
\begin{equation}
\label{rdp_loss}
    L_{rdp} = L_{rd} + \sum_{i=1}^{N} \lambda_{per} \|\mathcal{E}(\hat{x}_i)-\mathcal{E}(x_i)\|_2 + \lambda_{sty}\|G(\mathcal{E}(\hat{x}_i))- G(\mathcal{E}(x_i))\|_1,
\end{equation}
where $\mathcal{E}(\cdot)$ denotes the pretrained VGG network and $G(\cdot)$ is the Gram matrix of the given feature.

\section{Experiments}
\label{experiments}
\subsection{Experimental Settings}
\subsubsection{Datasets}
We train the proposed method on two widely used datasets: Vimeo-90k \cite{Vimeo90k} and REDS \cite{REDS}. Vimeo-90k contains 64,612 video clips, each composed of seven consecutive frames, while REDS consists of 262 videos with 100 frames each, following the protocol of \cite{BasicVSR_CVPR_2021}.

For evaluation, we adopt three standard benchmarks: REDS4 \cite{EDVR_CVPRW}, which includes four sequences at a resolution of 1280$\times$720; UVG \cite{UVG}, comprising seven 1080p videos; and MCL-JCV \cite{MCL_JCV}, containing 30 diverse 1080p video sequences. Following prior studies \cite{DCVC_HEM_ACMMM2022, DCVC_DC_CVPR2023, DCVC_FM_CVPR2024}, we evaluate the first 96 frames of each sequence and set the group of pictures (GOP) size to 8.

\subsubsection{Evaluation Metrics}
To evaluate the perceptual quality of reconstructed frames, we employ perceptual fidelity metrics, including \textit{LPIPS} \cite{LPIPS} and \textit{DISTS} \cite{DISTS}. We also report distortion-oriented metrics, namely \textit{PSNR} and \textit{MS-SSIM} \cite{MSSSIM}, to assess pixel-level reconstruction accuracy. In addition, temporal consistency is evaluated using \textit{tLPIPS} \cite{TCM_TOG}. The bitrate is measured in bits per pixel (\textit{bpp}).

\subsubsection{Implementation Details}
We adopt a two-stage training strategy for the proposed video compression framework. In the first stage, the model is trained on the REDS dataset using the Adam optimizer \cite{Adam} with a fixed learning rate of 10$^{-4}$ for 60 epochs. The batch size and GOP size are set to 4 and 5, respectively. Different bitrate models are obtained by varying the rate–distortion tradeoff parameter $\lambda \in \left\{128, 256, 512\right\}$ in Eq.~(\ref{rd_loss}). During training, video frames are randomly cropped into 256$\times$256 patches.

In the second stage, we fine-tune the model on the Vimeo-90k dataset for 50 epochs using Eq.~(\ref{rdp_loss}). The perceptual and style weights are set to $\lambda_{per}=1.0$ and $\lambda_{sty}=0.15$, respectively. The batch size and GOP size are set to 2 and 7, respectively. We use Adam with an initial learning rate of 10$^{-4}$, which is halved at the 45th and 48th epochs. All experiments are conducted on a single NVIDIA GeForce RTX 4090 GPU.
\subsection{Comparisons with State-of-the-Art Methods}
We compare our method against several state-of-the-art approaches, including ICISP \cite{ICISP_NN2025}, DCVC \cite{DCVC_NeurIPS2021}, DCVC-HEM \cite{DCVC_HEM_ACMMM2022}, DCVC-DC \cite{DCVC_DC_CVPR2023}, DCVC-FM \cite{DCVC_FM_CVPR2024}, Distributed-DVC \cite{Distributed_DVC_ICME2023}, DHVC \cite{DHVC_AAAI2024}, and GLC-video \cite{GLC_Video}.
Note that ICISP is a recently proposed perceptual image compression method that delivers strong perceptual quality.
\begin{figure}[htbp]
\centering
\includegraphics[width=1.0\textwidth]{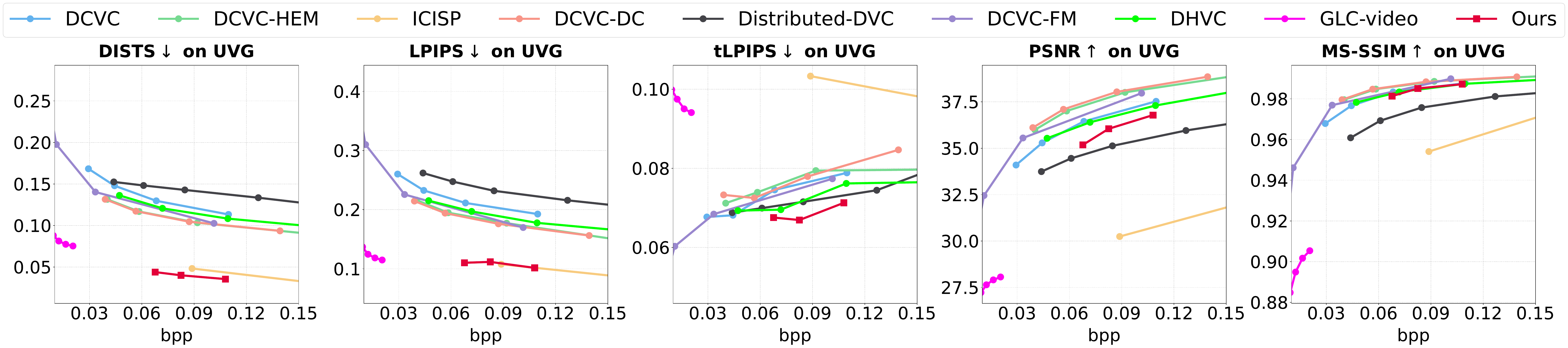}
\includegraphics[width=1.0\textwidth]{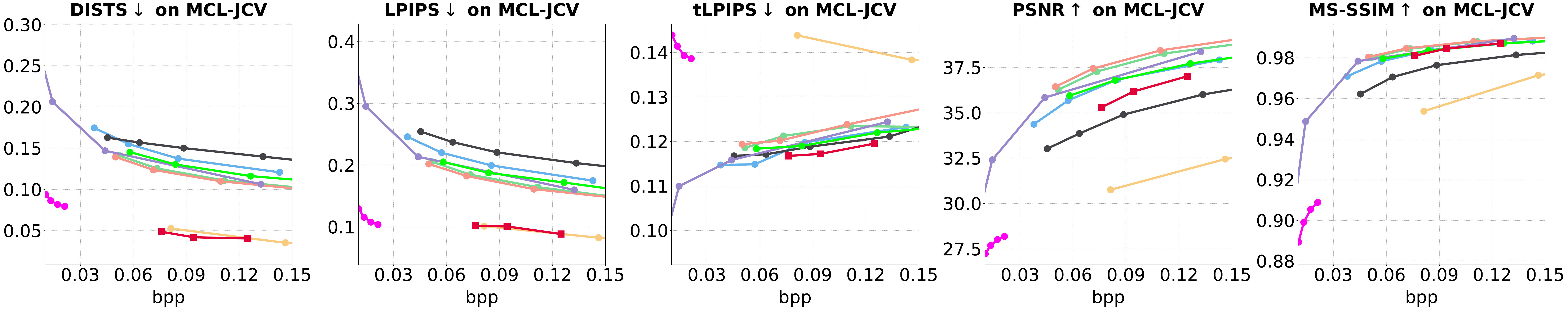}
\includegraphics[width=1.0\textwidth]{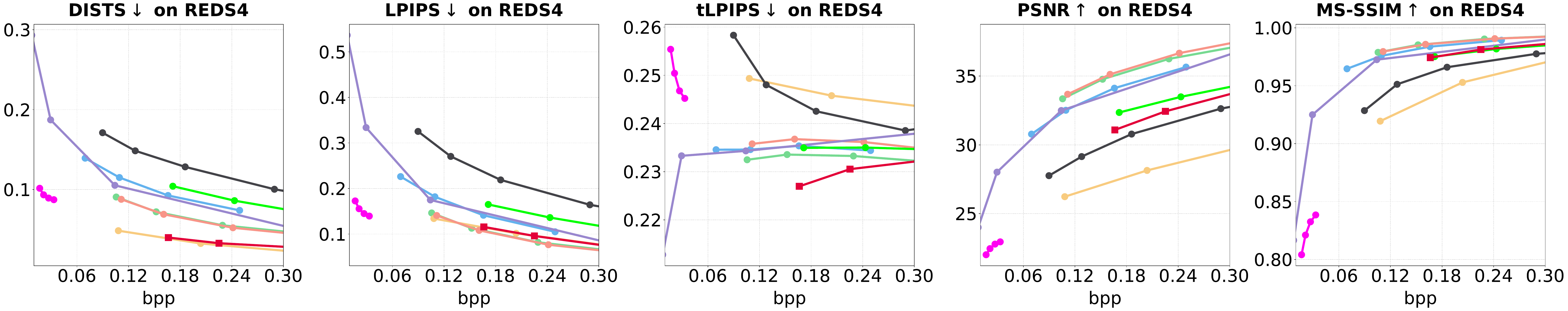}\vspace{-3mm}
\caption{Quantitative comparisons with state-of-the-art methods on benchmark datasets.} 
\label{quan_com}
\end{figure}

Fig. \ref{quan_com} presents quantitative comparisons across the benchmark datasets. The proposed method consistently outperforms hybrid video coding approaches (e.g., the DCVC series), Distributed-DVC, and DHVC in terms of perceptual metrics, including DISTS and LPIPS. Although ICISP achieves comparable perceptual quality, it does so at the expense of pixel-level fidelity, as reflected by its substantially lower PSNR and MS-SSIM scores compared to our method. Furthermore, our approach attains the best temporal consistency, achieving the lowest tLPIPS value among all competing methods.

\begin{figure*}[!t]
\scriptsize
\centering
    \begin{tabular}{c c c c}
            \multicolumn{2}{c}{\includegraphics[width=0.485\linewidth,height=0.3\linewidth]{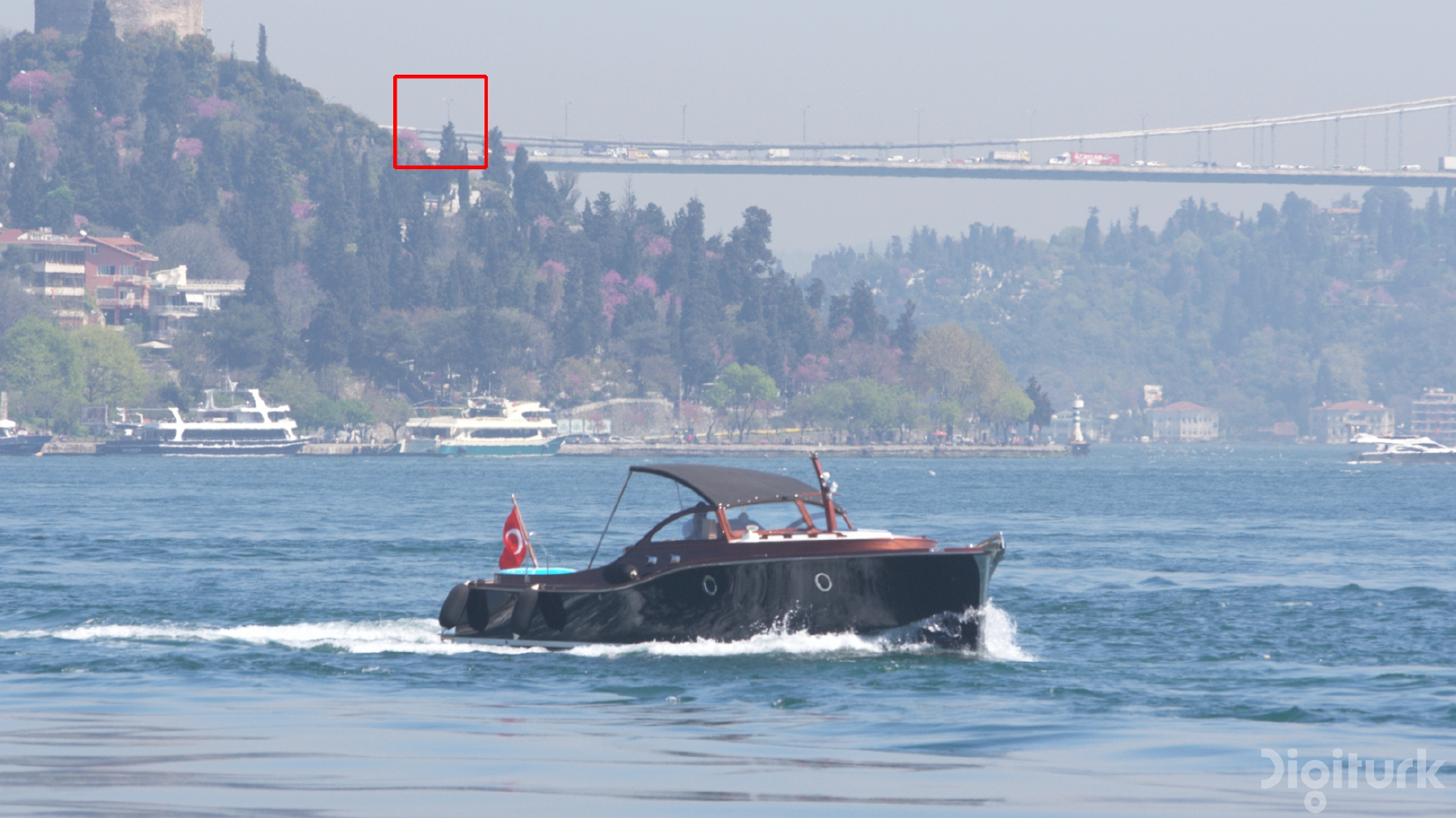}}
            \hspace{-3mm}& \multicolumn{2}{c}{\includegraphics[width=0.485\linewidth,height=0.3\linewidth]{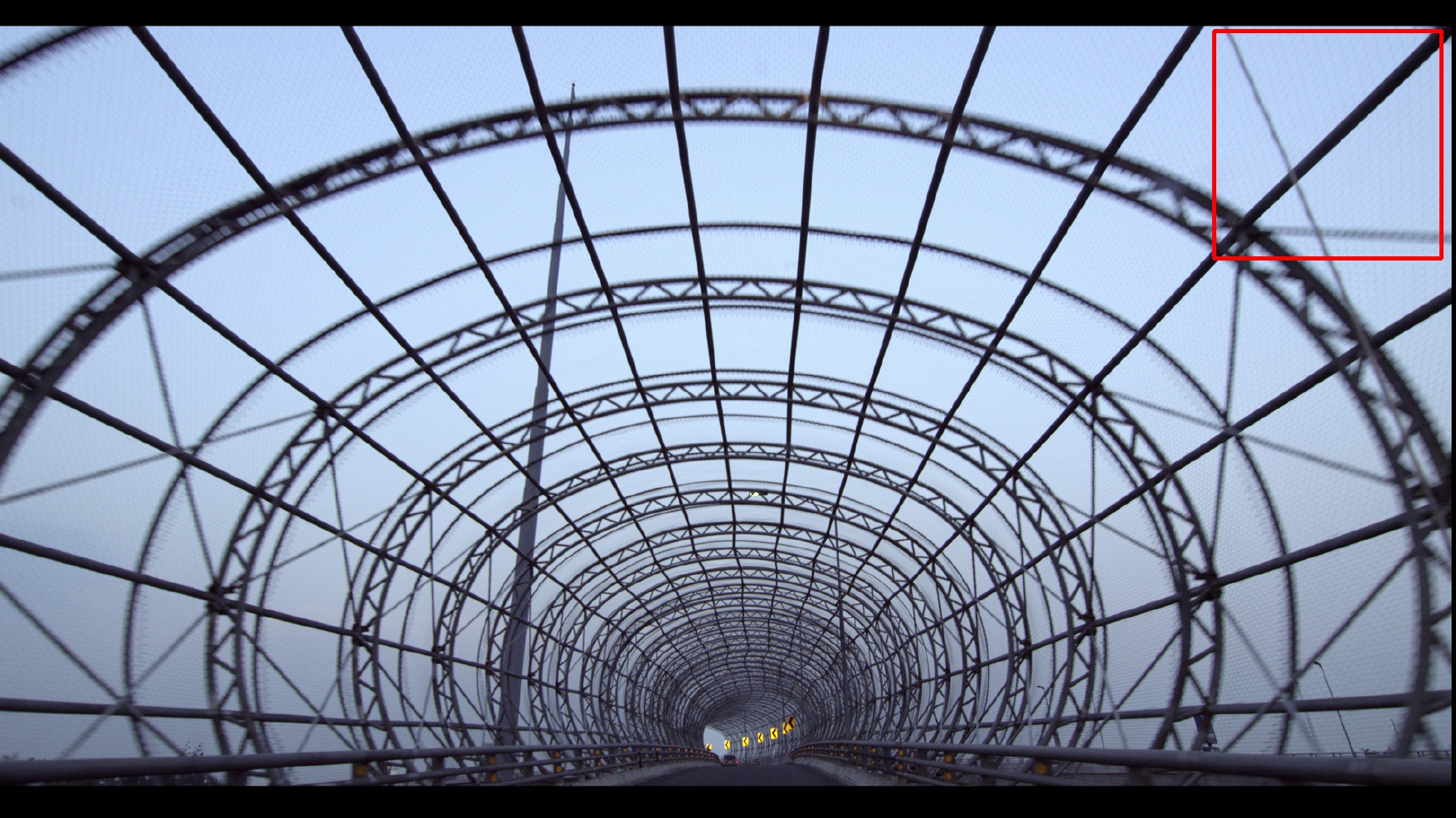}} \\
            \makebox[0.48\linewidth]{\textit{Frame 00000008, Clip Bosphorus}} 
            \makebox[0.48\linewidth]{\textit{Frame 00000016, Clip videoSRC10\_1920$\times$1080\_30}}  \\
            \includegraphics[width=0.24\linewidth,height=0.16\linewidth]{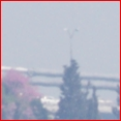}
            \includegraphics[width=0.24\linewidth,height=0.16\linewidth]{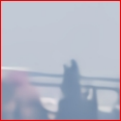}
            \includegraphics[width=0.24\linewidth,height=0.16\linewidth]{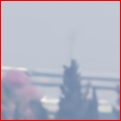}
            \includegraphics[width=0.24\linewidth,height=0.16\linewidth]{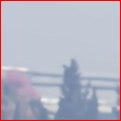} \\
            \makebox[0.24\linewidth]{Original patch (24)}
            \makebox[0.24\linewidth]{DCVC (0.0130)}
            \makebox[0.24\linewidth]{DCVC-DC (0.0120)}
            \makebox[0.24\linewidth]{DCVC-FM (0.0126)} \\		
            \includegraphics[width=0.24\linewidth,height=0.16\linewidth]{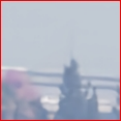}
            \includegraphics[width=0.24\linewidth,height=0.16\linewidth]{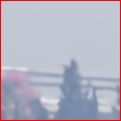}
            \includegraphics[width=0.24\linewidth,height=0.16\linewidth]{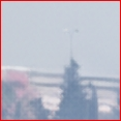}
            \includegraphics[width=0.24\linewidth,height=0.16\linewidth]{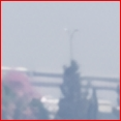}
            \\
            \makebox[0.24\linewidth]{Distributed-DVC (0.0195)}
            \makebox[0.24\linewidth]{DHVC (0.0143)}
            \makebox[0.24\linewidth]{ICISP (0.0813)}
            \makebox[0.24\linewidth]{Ours (\textbf{0.0080})}\\
             \includegraphics[width=0.24\linewidth,height=0.16\linewidth]{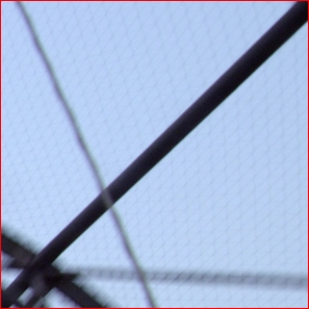}
            \includegraphics[width=0.24\linewidth,height=0.16\linewidth]{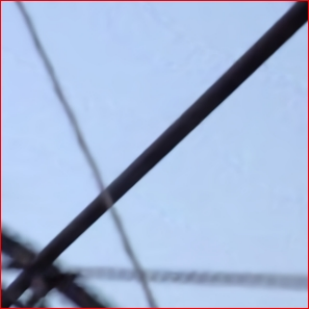}
            \includegraphics[width=0.24\linewidth,height=0.16\linewidth]{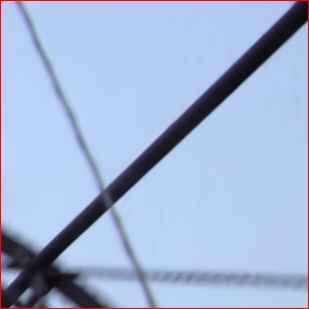}
            \includegraphics[width=0.24\linewidth,height=0.16\linewidth]{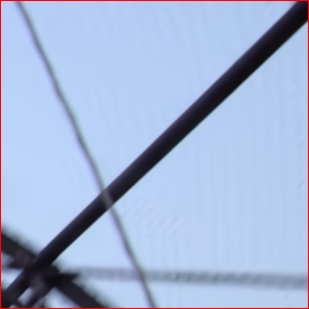} \\
            \makebox[0.24\linewidth]{Original patch (24)}
            \makebox[0.24\linewidth]{DCVC (0.1139)}
            \makebox[0.24\linewidth]{DCVC-DC (0.1281)}
            \makebox[0.24\linewidth]{DCVC-FM (0.1761)} \\		
            \includegraphics[width=0.24\linewidth,height=0.16\linewidth]{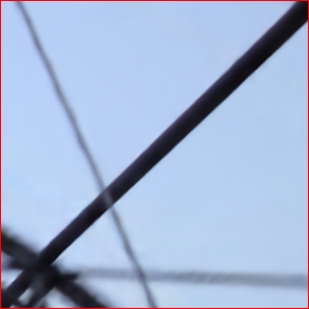}
            \includegraphics[width=0.24\linewidth,height=0.16\linewidth]{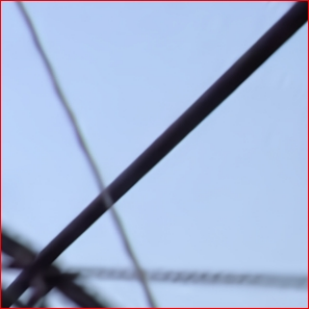}
            \includegraphics[width=0.24\linewidth,height=0.16\linewidth]{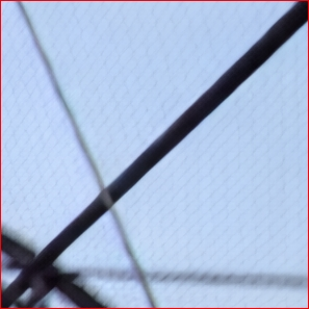}
            \includegraphics[width=0.24\linewidth,height=0.16\linewidth]{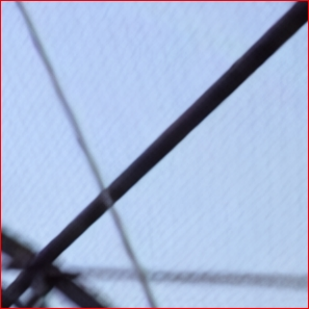}
            \\
            \makebox[0.24\linewidth]{Distributed-DVC (0.1197)}
            \makebox[0.24\linewidth]{DHVC (0.1384)}
            \makebox[0.24\linewidth]{ICISP (0.1584)}
            \makebox[0.24\linewidth]{Ours (\textbf{0.1044})}\\
    \end{tabular}
\vspace{-3mm}
\caption{Visual comparisons of different compression methods. The number in parentheses denotes the bpp used for compression. Compared with competing approaches, our method produces reconstructions with more faithful structural details at lower bitrates.}
\label{vis_com}
\end{figure*}

The visual comparison results are shown in Fig.\ref{vis_com}. Competing methods tend to produce either overly smooth or visually unrealistic reconstructions, often with higher bitrate consumption. In contrast, our method yields more visually pleasing results with better-preserved structural details at lower bitrates, as evidenced by the faithful reconstruction of structures such as the streetlight and the bridge.
%-------------------------------------------------------------------------
%-------------------------------------------------------------------------

\subsection{Analysis and Discussion}
\begin{figure*}
\scriptsize
\centering
\includegraphics[width=0.325\linewidth,height=0.325\linewidth]{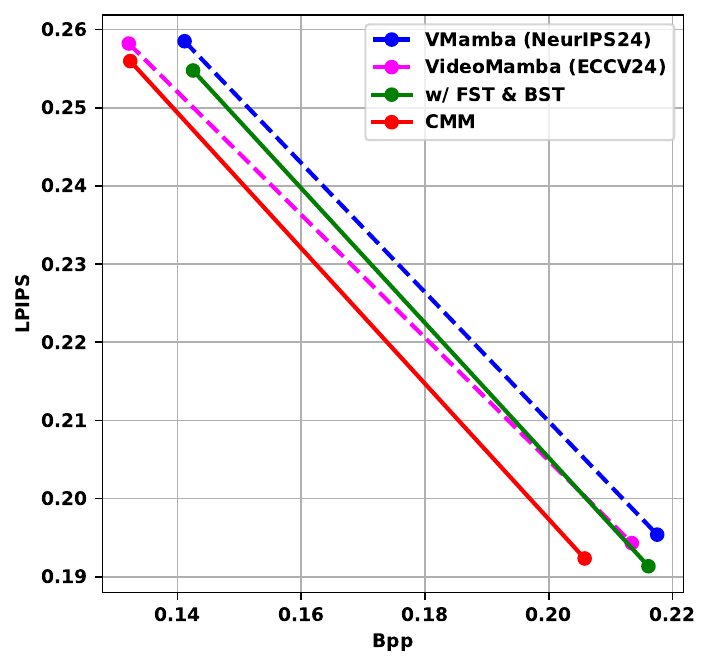}
\includegraphics[width=0.325\linewidth,height=0.325\linewidth]{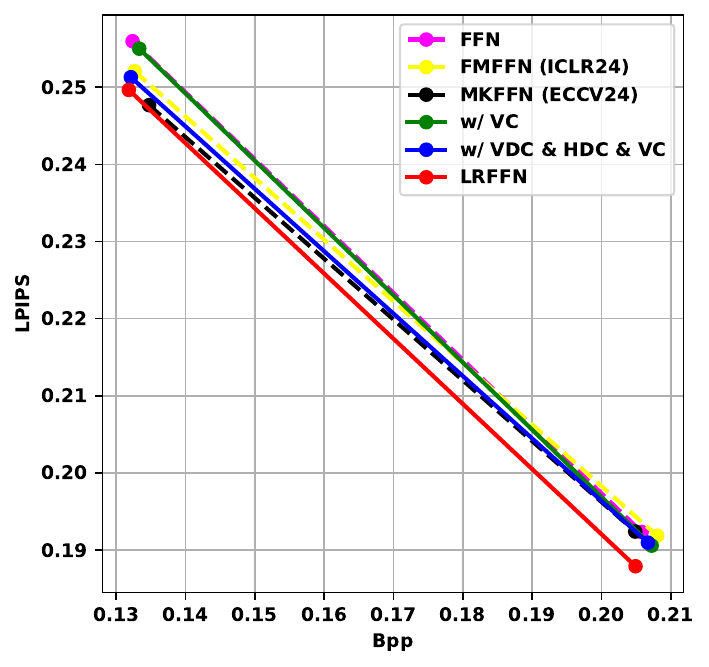} 
\includegraphics[width=0.325\linewidth,height=0.325\linewidth]{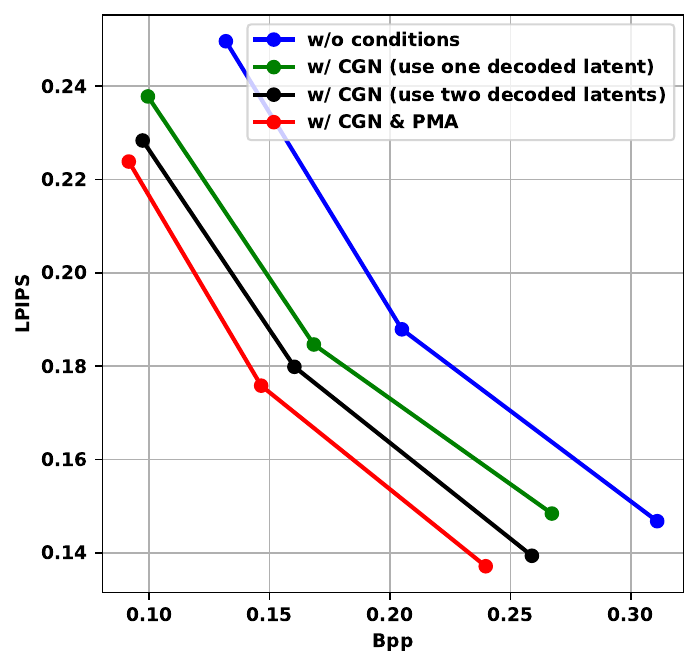} \\
\makebox[0.325\linewidth]{(a)} 
\makebox[0.325\linewidth]{(b)}
\makebox[0.325\linewidth]{(c)} \\
\vspace{-3mm}
\caption{Quantitative comparison of individual components under the rate–perception tradeoff.} 
\label{rp_curve}
\end{figure*}
\subsubsection{Effectiveness of CMM}
%-------------------------------------------------------------------------
\begin{figure*}[!t]
\scriptsize
\centering
\begin{tabular}{ccc}
    \includegraphics[width=0.25\linewidth]{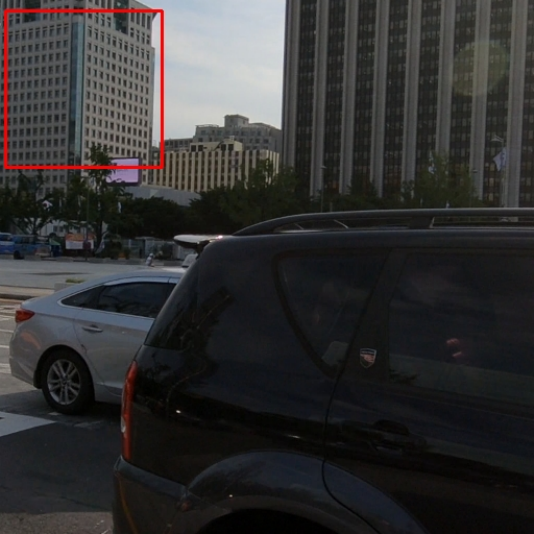} 
    \includegraphics[width=0.25\linewidth]{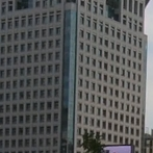} 
    \includegraphics[width=0.25\linewidth]{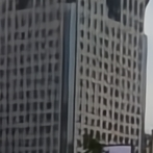} \\
    \makebox[0.25\linewidth]{\textit{Frame 009, Clip 015}} 
    \makebox[0.25\linewidth]{(a) Original patch (24)} 
    \makebox[0.25\linewidth]{(b) VMamba (0.0758)} \\
    \includegraphics[width=0.25\linewidth]{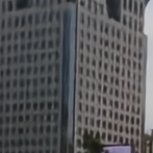} 
    \includegraphics[width=0.25\linewidth]{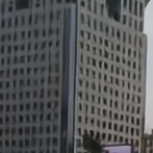} 
    \includegraphics[width=0.25\linewidth]{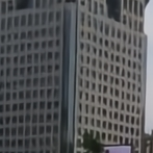} \\
    \makebox[0.25\linewidth]{(c) VideoMamba (0.0697)} 
    \makebox[0.25\linewidth]{(d) w/ FST $\&$ BST (0.0764)} 
    \makebox[0.25\linewidth]{(e) CMM (\textbf{0.0682})} \\
\end{tabular}
\vspace{-3mm}
\caption{Effectiveness of the proposed CMM.}
\label{stm_com}
\end{figure*}
%-------------------------------------------------------------------------
The proposed CMM is designed to effectively capture both spatial and temporal long-range dependencies, thereby enhancing video compression performance. To evaluate its effectiveness, we remove CMM from the proposed method and retain only the bidirectional spatial scanning. As shown in Fig. \ref{rp_curve}(a), using CMM improves compression performance by leveraging a bidirectional temporal scanning strategy (red solid curve vs. green solid curve), demonstrating that scanning features along the temporal dimension is crucial for video compression. Fig. \ref{stm_com}(d) and (e) further indicate that using the CMM can generate better reconstruction results at lower bitrates.

Furthermore, we compare our approach with related methods, including VMamba \cite{VMamba} and VideoMamba \cite{VideoMamba_ECCV}. All compression networks are trained from scratch using the same experimental settings. 
Fig. \ref{rp_curve}(a) demonstrates that both VMamba and VideoMamba fall short in achieving optimal performance: VMamba applies only a cross-scan along the spatial dimension, while VideoMamba employs a spatial-first bidirectional scan. These limitations suggest that neither method fully exploits the temporal redundancies across video frames. Fig. \ref{stm_com}(b) and (c) show that VMamba and VideoMamba cannot produce better decoded frames, while the proposed CMM reconstructs clearer frames at lower bitrates (Fig. \ref{stm_com}(e)).

\subsubsection{Impact of LRFFN}
The LRFFN is designed to capture local dependencies by integrating the proposed hybrid convolution block. To evaluate its effectiveness, we first compare it with several variants, including the vanilla feed-forward network (FFN), the frequency-modulation feed-forward network (FMFFN) \cite{FTIC_ICLR2024}, and the multi-kernel ConvFFN (MKFFN) \cite{MKFFN_ECCV2025}. 

\begin{figure*}[!t]
\scriptsize
\centering
\begin{tabular}{cccccc}
    \includegraphics[width=0.19\linewidth]{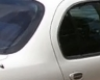} 
    \includegraphics[width=0.19\linewidth]{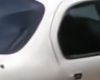} 
    \includegraphics[width=0.19\linewidth]{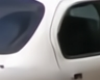} 
    \includegraphics[width=0.19\linewidth]{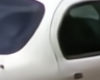} 
    \includegraphics[width=0.19\linewidth]{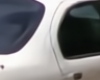} \\
    \makebox[0.19\linewidth]{Original patch (24)} 
    \makebox[0.19\linewidth]{FFN (0.1732)} 
    \makebox[0.19\linewidth]{FMFFN (0.1728)}
    \makebox[0.19\linewidth]{MKFFN (0.1760)}
    \makebox[0.19\linewidth]{LRFFN (\textbf{0.1692})}
\end{tabular}
\vspace{-3mm}
\caption{Visual comparisons of different FFN variants for video compression.}
\label{lrffn_com}
\end{figure*}

As shown in Fig. \ref{rp_curve}(b), our LRFFN consistently outperforms these alternatives in compression performance, demonstrating a stronger ability to leverage local context and preserve fine-grained details. Visual comparisons in Fig.\ref{lrffn_com} further reveal that LRFFN reconstructs higher-quality frames at lower bitrates, with more accurate structural restoration—for instance, the structure of the car is clearly preserved.

We further conduct ablation studies, as shown in Fig. \ref{rp_curve}(b), to validate the contribution of each component within our LRFFN. Incorporating vertical and horizontal difference convolutions results in improved compression performance (blue curve vs. green curve), demonstrating the effectiveness of difference convolutions in video compression. Furthermore, introducing the proposed hybrid convolution block for local feature refinement enables LRFFN to surpass all other variants (see red curve). 
%Visual comparisons also confirm that LRFFN produces significantly higher-quality frames at lower bitrates.
%-------------------------------------------------------------------------
%-------------------------------------------------------------------------

\subsubsection{Effect of CCEM}
The conditional channel-wise entropy model (CCEM) is designed to exploit temporal redundancy in latent representations by leveraging effective condition priors. These priors, generated by the PMA and CGN, enable a more accurate estimation of the conditional probability for the current latent encoding, thereby minimizing entropy and maximizing coding efficiency. To assess the impact of these condition priors, we compare our method with a baseline that does not use any conditions, training it from scratch for fair comparisons.

As shown in Fig. \ref{rp_curve}(c), using previously decoded latent features, i.e., $\bar{\text{y}}_{t-2}$ and $\bar{\text{y}}_{t-1}$, as conditions significantly improves compression performance compared to the baseline. Moreover, incorporating pseudo-aligned features as supplementary conditions boosts performance even more (red curve vs. black curve). 
%Visual comparisons in Fig. \ref{qua_ccem} also demonstrate that the CCEM generates better results at lower bitrates. 

\subsubsection{Temporal Consistency}
\begin{figure}
\centering
\includegraphics[width=1.0\textwidth]{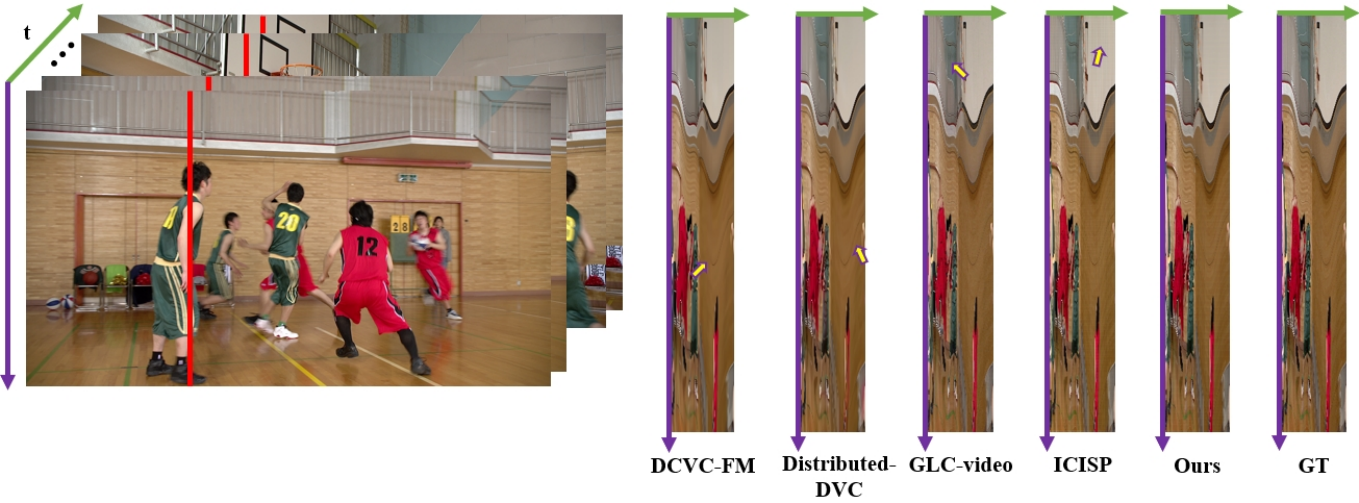}\vspace{-3mm}
\caption{Visual comparisons of temporal consistency for decoded frames. We visualize pixel values along selected columns, indicated by red solid lines in the original video frames. Zoom in for best view.} 
\label{tc_com}
\end{figure}
We further evaluate the temporal consistency of the decoded video frames by visualizing the changes across time. As shown in Fig. \ref{tc_com}, previous methods either exhibit obvious discontinuities or introduce artifacts. In contrast, our method generates videos with better temporal consistency properties.
\begin{table}[htbp]
\centering
\scriptsize
\caption{Model complexity comparison on the video frames with 512$\times$512 resolution. The encoding and decoding times are measured on a machine with an NVIDIA GeForce RTX 3090 GPU. Best and second-best results are marked in \textbf{bold} and \underline{underlined}, respectively.}
\begin{tabular}{llll}
\hline
  Methods & Parameters (M) & Encoding time (ms) & Decoding time (ms) \\ 
  \hline
  %ICISP & \underline{29.26} & 68.13 & 70.37 \\
  DCVC & \textbf{19.77} & 884.63  & 4,092,38 \\
  DCVC-HEM & 48.69 & 65.58 & \textbf{44.12} \\
  DCVC-DC & 50.79 & 57.49 & 90.54 \\
  DCVC-FM & 44.93 & \underline{53.97} & 131.60 \\
  %Distributed-DVC & 47.21 & $>$3.7 & $>$29.96 \\
  Distributed-DVC & \underline{47.21} & N/A & N/A \\
  DHVC & 117.92 & 59.53 & \underline{57.29} \\
  GLC-video & 289.33 & \textbf{15.59} & 61.10 \\
\hline
  Ours & 47.79 & 68.98 & 76.97 \\
 % Ours (w/o CCEM) & 2.53 & 28.58 & 24.04 \\
\hline
\end{tabular}
\label{model_complexity_com}
\end{table}
\subsubsection{Model Complexity Comparisons}
We compare the model complexity of our method with state-of-the-art compression methods in terms of the number of network parameters, encoding, and decoding time. Table~\ref{model_complexity_com} shows that our method contains fewer parameters than DHVC and GLC-Video, while exhibiting a parameter count comparable to that of most hybrid video coding methods. Furthermore, our method achieves faster encoding and decoding speeds than DCVC, but does not consistently outperform all competing methods. We conduct a component-wise analysis and find that the proposed conditional channel-wise entropy model comprises 45.26M parameters and incurs encoding and decoding times of 40.40 ms and 52.93 ms, respectively, dominating both the parameter count and inference time. Future research will investigate more efficient entropy models designed for transform-based video compression.

%our method does not outperform competing approaches in terms of model complexity. We observe that the proposed conditional channel-wise entropy model accounts for the largest proportion of network parameters and inference time (see ``Ours (w/o CCEM)"). Future work will explore more efficient entropy models tailored for transform-based video compression.

\section{Conclusion}
\label{conclusion}
In this paper, we propose a simple yet effective transform-based perceptual video compression approach that does not involve complex explicit motion estimation and compensation operations.
We introduce a cascaded Mamba module augmented by geometric transformations to capture long-range spatial and temporal dependencies and develop a locality refinement feed-forward network to enable local modeling. 
%These components are integrated into the encoder and decoder of the video compression framework to facilitate compact feature representation.
%
Additionally, we propose a conditional channel-wise entropy model that enhances current frame coding by accurately estimating the probabilities using more effective conditional priors. 
Both quantitative and qualitative results demonstrate the effectiveness of our method in compressing videos at low bitrates.

\section{Acknowledgements}
This work are supported by the National Natural Science Foundation of China (NSFC62376208, U24A20265), Key Core Technology Research Project in Shaanxi Province (2024SF-GJHX-27) and Qin Chuangyuan General Window "Four Chain" Integration Project (2024PT-ZCK-66).

\bibliographystyle{elsarticle-num} 
\bibliography{reference}

%\begin{thebibliography}{00}

%% For numbered reference style

%\bibitem{lamport94}
  %Leslie Lamport,
  %\textit{\LaTeX: a document preparation system},
  %Addison Wesley, Massachusetts,
 % 2nd edition,
 % 1994.

%\end{thebibliography}
\end{document}